\documentclass[conference]{IEEEtran}
\IEEEoverridecommandlockouts
\usepackage[utf8]{inputenc}
\usepackage[T1]{fontenc}
\usepackage{graphicx}
\usepackage{amsmath,amssymb}
\usepackage{booktabs}
\usepackage{xcolor}
\usepackage{url}

\graphicspath{{figures/}}
\newcommand{\unit}[1]{\,\mathrm{#1}}

\title{A Hardware-Partitioned micro-ROS Mini-Vehicle for Operator-Supervised
Neural Driving}

\author{\IEEEauthorblockN{Gustavo Claudio Karl Couto, Eric Aislan Antonelo
and Gabriel George Zipperer}
\IEEEauthorblockA{Automation and Systems Engineering Department\\
Federal University of Santa Catarina, Florian\'opolis, Brazil\\
gustavo.karl.couto@posgrad.ufsc.br, eric.antonelo@ufsc.br}\thanks{\copyright~2026 IEEE. Personal use of this material is permitted. Permission from IEEE must be obtained for all other uses, in any current or future media, including reprinting/republishing this material for advertising or promotional purposes, creating new collective works, for resale or redistribution to servers or lists, or reuse of any copyrighted component of this work in other works. Accepted for publication in the XVI Brazilian Symposium on Computing Systems Engineering (SBESC 2026).}}

\begin{document}
\maketitle

% -----------------------------------------------------------------------------
% Abstract (DeepPicar pattern): present the platform -> what we characterize ->
% findings paragraph.
% -----------------------------------------------------------------------------
\begin{abstract}
We present a low-cost miniature Ackermann vehicle for embedded neural-driving
research, organized as a hardware-partitioned perceive--decide--act agent: an
ESP32 camera node streams frames over Wi-Fi to a host computer, where neural
inference and state estimation run, while an ESP32-S3 chassis controller
executes safety-critical actuation in a $100\unit{Hz}$ per-wheel proportional--integral--derivative (PID) loop
beneath a micro-ROS client. The partition isolates real-time control and safety on
the vehicle while learning-based components run off-board, supervised by an
operator from a gamepad. We characterize the complete pipeline during real
closed-loop neural driving. Wireless sensing, rather than neural inference,
dominates latency, with a $59.5\unit{ms}$ median capture-to-actuation delay;
the platform absorbed wireless variability by holding the last safe command
under a firmware watchdog, operating across twelve autonomous sessions with
zero collisions or manual interventions; and encoder--inertial-measurement-unit (IMU) fusion cut odometry loop-closure
error by $22\%$. These results indicate that carefully partitioned
microcontroller-class vehicles support reproducible experiments in embedded
AI, wireless control, and operator-supervised autonomous driving. The
platform software is open source.
\end{abstract}


% =============================================================================
\section{Introduction}
\label{sec:intro}
% P1 (DeepPicar P1): context — embedded AI workloads in AVs.
% =============================================================================
Deep neural networks have become a standard component of autonomous-vehicle
software, including end-to-end policies that map camera images directly to
actuation commands~\cite{bojarski2016,couto2023hgail}. Executing these
workloads on physical vehicles imposes requirements that pure simulation
studies never exercise: the perception-to-actuation pipeline must sustain a
sufficient control rate, sensing latency must remain within the tolerance of
the control loop, and the platform must fail safely when any link of the
pipeline stalls.

% P2 (DeepPicar P2): the conflicting requirements.
On low-cost research platforms these requirements conflict with the available
resources. Microcontroller-class boards cost a few dollars and consume little
power, but cannot run modern neural networks; single-board computers can, at
higher cost and power and with a model-compression stage; and offloading over a
wireless link adds transport latency and variability to the control loop. The
platform designer must therefore decide \emph{where} each function
runs---low-level control, sensing, state estimation, inference---and quantify
the timing consequences of that partition.

% P3 (DeepPicar P3): need for testbeds + our prior platform work.
Low-cost testbeds are the established vehicle for studying these trade-offs:
DeepPicar characterized the real-time behavior of DAVE-2-class inference on a
Raspberry Pi~\cite{bechtel2018deeppicar}, and platforms such as F1TENTH and
Duckietown support systematic experimentation at the \$$300$--\$$3000$
scale~\cite{okelly2020f1tenth,paull2017duckietown}. In previous work we
introduced a miniature Ackermann platform for imitation-learning research,
with a printed urban track and map-referenced trajectory
evaluation~\cite{couto2026platform}; that work reported the embedded
system only as far as needed to reproduce its learning results.

% P4: what this paper is.
This paper presents the platform from the systems perspective. The design
splits on-board computing between \emph{two} microcontrollers---an ESP32-S3
dedicated to real-time chassis control, closing a $100\unit{Hz}$ per-wheel
proportional--integral--derivative (PID) loop that acts as the vehicle's electronic differential, under an ESP-IDF
firmware with a micro-ROS client~\cite{belsare2023micro}, and an ESP32
camera module dedicated to sensing---while neural inference and sensor fusion
run on a host computer over Wi-Fi. The partition is the classic
perceive--decide--act decomposition of an embodied agent mapped onto
hardware: the camera node is the perception module, the host the decision
module, and the chassis controller the action module. It keeps the two
microcontroller nodes under \$$80$ and the complete vehicle near \$$235$
(Table~\ref{tab:bom}), isolates the safety-critical $100\unit{Hz}$ actuation
loop from all best-effort traffic, and makes the platform's real-time
behavior a measurable, three-hop pipeline.

% P5 (DeepPicar P5): research questions.
Using data from twelve real closed-loop neural driving sessions, we address
three questions: (1)~what latency and effective throughput does the
camera-to-actuation pipeline actually deliver, compared to its configured
rates; (2)~how accurate is the on-board state estimation that the firmware
and host derive from encoders, chassis geometry, and the inertial unit; and
(3)~how does the safety layer behave when the pipeline degrades---frames are
delayed, dropped, or the wireless link stalls.

% P6: findings preview (honest).
Our main observations: the wireless camera link, not inference, dominates
a $60\unit{ms}$ median capture-to-actuation budget; wireless delivery is
variable, with occasional second-long frame gaps, yet the watchdog and
command-gating layer turned every such event into safe coasting, with zero
collisions or interventions; and heading, not translation, dominates
odometry drift, so fusing encoder translation with gyroscope heading cuts
loop-closure error by $22\%$.

The main contributions are:
\begin{enumerate}
  \item the design and rationale of a dual-microcontroller micro-ROS
        architecture that isolates hard-periodic actuation from best-effort
        sensing and offloaded inference, supervised in real time from the
        operator's gamepad;
  \item a timing characterization of the full sensing-to-actuation pipeline
        under real neural closed-loop operation, rather than synthetic load;
  \item an empirical comparison of four on-board odometry estimators and of
        the safety layer's behavior under pipeline degradation, with the
        resulting design lessons.
\end{enumerate}
The firmware for both microcontrollers, the host software, and the deployed
policy checkpoint are released under the MIT
license.\footnote{\url{https://github.com/gustavokcouto/teledriving-in-miniature}}

% =============================================================================
\section{Background and Related Work}
\label{sec:related}
% =============================================================================
\subsection{Low-cost autonomous-vehicle testbeds}
DeepPicar~\cite{bechtel2018deeppicar} replicates NVIDIA's DAVE-2 network on a
\$$70$ Raspberry Pi vehicle and remains the reference study of real-time convolutional neural network (CNN)
inference on low-cost hardware; its analysis is centered on the computing
platform (execution times, contention, isolation). F1TENTH targets
$1/10$-scale racing with LiDAR and a Jetson-class computer at a much higher
price point~\cite{okelly2020f1tenth}, and Duckietown scales classroom fleets
of camera-only robots~\cite{paull2017duckietown}. Our platform differs in its
computing partition---no on-board single-board computer at all, but two
microcontrollers plus an off-board host---and in that its evaluation
infrastructure (map-referenced trajectories on a printed
track~\cite{couto2026platform}) allows closing the loop between timing
behavior and control quality. This paper provides for our platform what
DeepPicar provided for the Pi: the systems-level characterization, measured,
however, during real autonomous driving sessions rather than under synthetic
workloads. The comparison is architectural: an on-board pipeline pays
inference but no transport, ours the reverse.

\subsection{Microcontroller-class robotics and micro-ROS}
micro-ROS brings the ROS\,2 node model, typed topics, and Data Distribution Service (DDS) transport
to microcontrollers~\cite{belsare2023micro}, allowing a
microcontroller-unit (MCU) firmware to participate in the same computation graph as host-side
nodes. Reported uses often concentrate on peripheral sensor nodes; here the
micro-ROS client is the vehicle's \emph{primary controller}, publishing
odometry and inertial-measurement-unit (IMU) streams at $20\unit{Hz}$ while consuming actuation
commands, with the hard $100\unit{Hz}$ control loop running beneath the
middleware inside the same firmware.

% =============================================================================
\section{Platform Architecture}
\label{sec:platform}
% =============================================================================
\begin{figure}[t]
  \centering
  \includegraphics[width=0.52\linewidth]{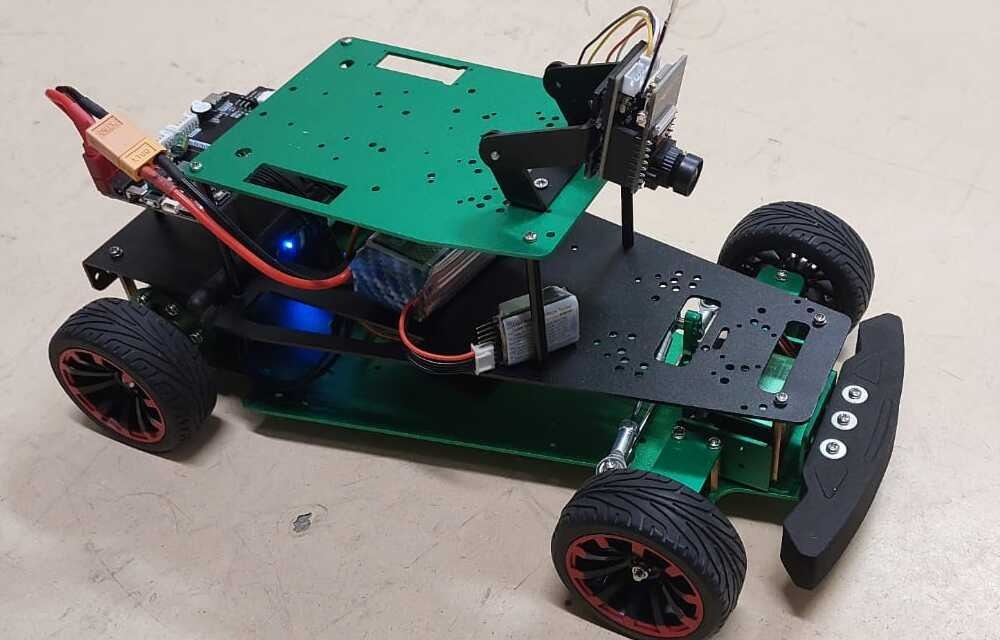}
  \caption{The vehicle: an off-the-shelf Ackermann chassis carrying two
  microcontrollers---an ESP32-S3 chassis controller and an ESP32 camera
  module on a short mast. There is no on-board single-board computer.}
  \label{fig:vehicle}
\end{figure}

\begin{figure*}[t]
  \centering
  \includegraphics[width=\textwidth]{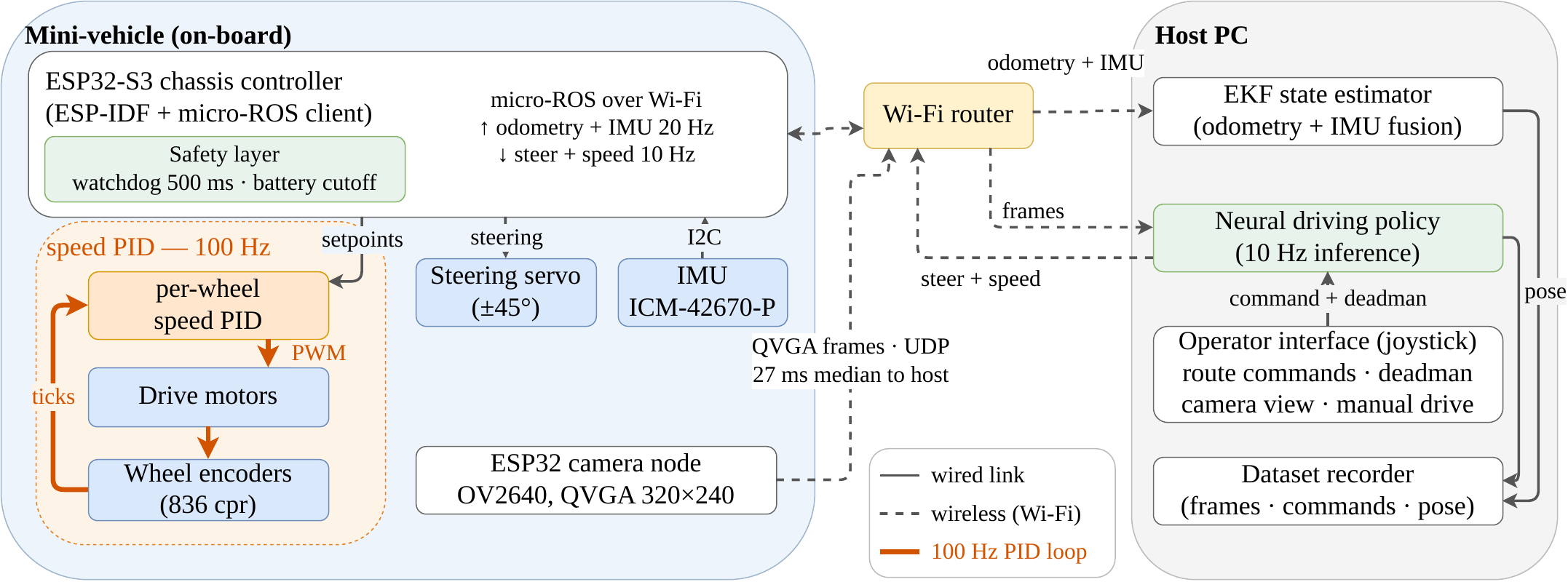}
  \caption{Hardware/software architecture: the ESP32-S3 chassis controller,
  the ESP32 camera node, the Wi-Fi link, the host-side estimator, policy,
  and operator interface. Solid edges: wired links; dashed: across Wi-Fi;
  thick orange: the $100\unit{Hz}$ control loop.}
  \label{fig:arch}
\end{figure*}

\begin{table}[t]
  \caption{Approximate bill of materials for the miniature vehicle
  platform.}
  \label{tab:bom}
  \centering
  \begin{tabular}{@{}lr@{}}
    \toprule
    Component & approx.\ US\$ \\
    \midrule
    Ackermann aluminum chassis (incl.\ $2\times$ MD520 & \\
    \quad encoder motors and steering servo)$^\dagger$ & $108$ \\
    micro-ROS control board (ESP32-S3, IMU) & $64$ \\
    ESP32 Wi-Fi camera module (OV2640) & $15$ \\
    LiPo battery 3S $11.1\unit{V}$ $2200\unit{mAh}$$^\dagger$ & $47$ \\
    \midrule
    \textbf{Vehicle total} & $\mathbf{\approx 234}$ \\
    \midrule
    Wi-Fi router (TP-Link TL-WR829N, infrastructure)$^\dagger$ & $16$ \\
    Tri-mode wireless gamepad (infrastructure) & $39$ \\
    Printed track map $3.2{\times}2.8\unit{m}$ (infrastructure) & $252$ \\
    \bottomrule
  \end{tabular}

  \smallskip
  \parbox{0.9\linewidth}{\footnotesize Manufacturer list prices (July 2026);
  items marked $^\dagger$ purchased locally in Brazil, converted at
  $\approx$R\$5.5/US\$. The chassis bundle includes the two encoder drive
  motors and the steering servo. Infrastructure items and the host PC (a
  consumer desktop shared across experiments) are not part of the vehicle
  total.}
\end{table}

\subsection{Design rationale}
\begin{figure}[t]
  \centering
  \includegraphics[width=0.75\linewidth]{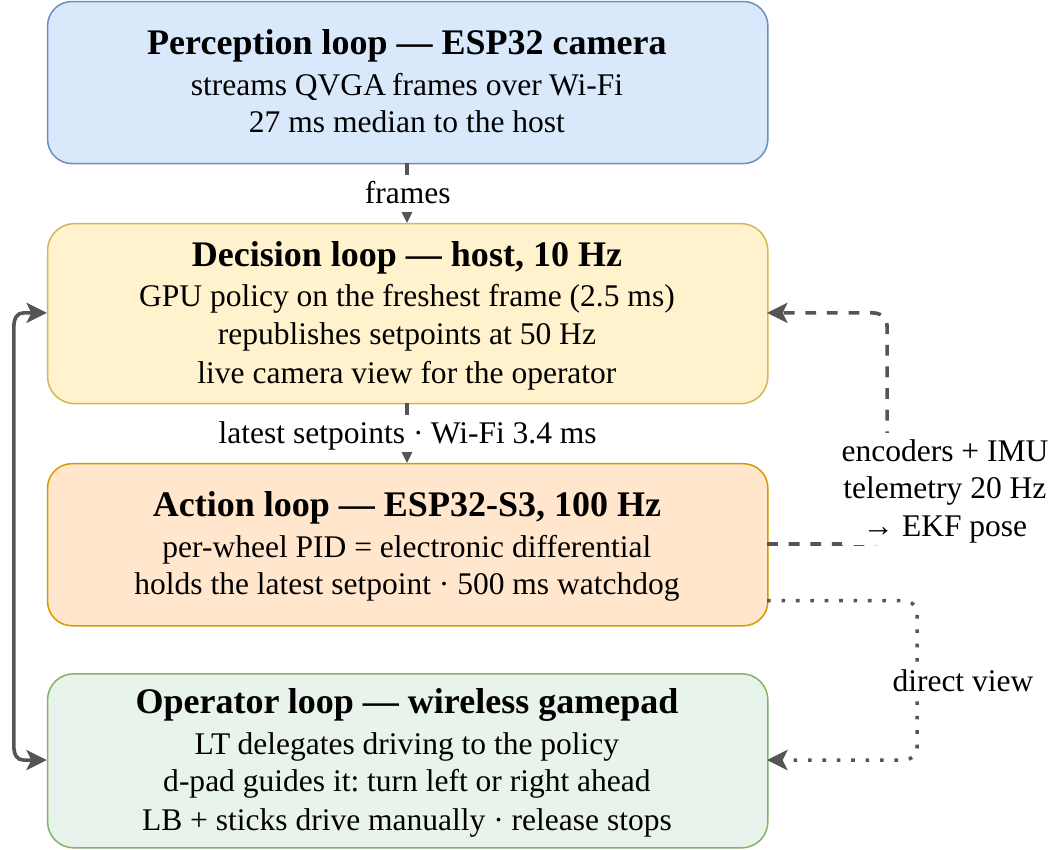}
  \caption{Functional perceive--decide--act view of the platform, one loop
  per module: camera streaming, host-side policy execution, chassis
  control, and operator supervision closing the loop.}
  \label{fig:loops}
\end{figure}

The architecture (Figs.~\ref{fig:vehicle} and~\ref{fig:arch}) follows one
principle: the actuation loop must not share a failure domain with anything
best-effort. Camera streaming, Wi-Fi transport, host inference, and even the
ROS middleware are all allowed to stall; the only component that must not
miss its period is the $100\unit{Hz}$ speed/steering loop, so it runs on its
own microcontroller, beneath (not inside) the micro-ROS executor, and is
guarded by mechanisms that degrade the vehicle to a safe stop whenever its
inputs go stale. Fig.~\ref{fig:loops} summarizes the resulting functional
cascade, one loop per module: the perception loop on the camera node streams
the scene continuously; the decision loop on the host decides at
$10\unit{Hz}$ on the freshest frame and republishes the latest decision at
$50\unit{Hz}$; the action loop on the chassis controller actuates at
$100\unit{Hz}$---each stage holds the latest output of the one above it, so
a perception hiccup never blocks actuation. Splitting the camera onto a second, independent ESP32
removes the streaming load---the single largest and burstiest producer---from
the controller's CPU and radio entirely.

A single microcontroller could host both loops under a scheduler that reserves processor time for the critical task (server-based or mixed-criticality scheduling~\cite{sprunt1989aperiodic,burns2017mixed}). We kept two chips because the off-the-shelf modules already come that way---the motor-driver board carries an ESP32-S3 without a camera interface, the camera module an ESP32 without motor hardware---and because on one chip the camera stream would compete with the control traffic for a single Wi-Fi radio, memory and power rail, not only for processor time; with two modules each has its own radio, and a stalled camera node cannot take actuation down with it. The controller's load headroom beside its $100\unit{Hz}$ loop, which would decide whether one chip suffices, was not measured.

Offloading to the host also removes the model-compression stage that on-board inference would impose---no quantizing or pruning to fit a single-board computer (SBC), so the usable models are bounded by the host graphics processing unit (GPU)---and lets policies, estimators and tooling iterate in standard Python/ROS~2 with no reflashing.

\subsection{Action module: chassis controller (ESP32-S3)}
The ESP-IDF firmware closes a per-wheel incremental PID speed loop at
$100\unit{Hz}$ (gains $0.75/0.12/0.04$) on $836$-count-per-revolution wheel
encoders and drives the steering servo, limited to $\pm45^\circ$ on the
$0.235\unit{m}$-wheelbase Ackermann geometry. The loop is per-wheel because
the two drive motors are independent---there is no mechanical differential:
for each command the firmware derives the inner and outer rear-wheel
setpoints from the Ackermann geometry as
$v\,(1 \mp \frac{W}{2L}\tan\delta)$ for track width $W$, wheelbase $L$, and
steering angle $\delta$, and regulates each wheel against its
own encoder, acting as an electronic differential. A single loop on the mean
speed would drag the inner wheel through tight turns, scrubbing speed
tracking and corrupting the per-wheel encoder odometry that
Sec.~\ref{sec:characterization} builds on. Speed commands are capped at
$0.7\unit{m/s}$ by the policy's output scaling and at $0.15\unit{m/s}$ in
autonomous operation. Two safety mechanisms are implemented in firmware, independent of the host:
a $500\unit{ms}$ command watchdog that zeroes the speed setpoint when no
fresh command arrives---the backstop for every host-side failure, including
a released deadman---and a battery-voltage cutoff. The
micro-ROS client publishes the integrated odometry twist and the raw IMU
(ICM-42670-P) at $20\unit{Hz}$ and subscribes to the actuation topic.

\subsection{Perception module: vision node (ESP32 camera)}
A separate ESP32 with an OV2640 sensor streams $320{\times}240$ (QVGA) JPEG frames
over a custom User Datagram Protocol (UDP) stream, chosen over the
Transmission Control Protocol (TCP) and RTSP to avoid retransmission
stalls: a late frame is a dropped frame, never a blocking one. The camera is
calibrated at QVGA from ChArUco chessboard corners~\cite{zhang2000flexible}.

\subsection{Decision module: host computer}
A commodity workstation (6-core/12-thread Ryzen~5~5500, $32\unit{GB}$ RAM,
NVIDIA RTX~3060 with $12\unit{GB}$ of VRAM) completes the graph with three
nodes: an extended
Kalman filter (EKF)~\cite{moore2016generalized} fusing the $20\unit{Hz}$ odometry
and IMU streams into the pose estimate; the neural driving policy, invoked at
$10\unit{Hz}$ on the most recent camera frame and publishing
steering/speed setpoints; and the operator's wireless gamepad, which
arbitrates control by strict priority: a brake button stops the vehicle
unconditionally; while the neural trigger (LT) is held, the policy
drives, guided live by the operator: the route command, picked on the
d-pad (\emph{follow-lane}, \emph{turn-left}, \emph{turn-right}), selects
its branch at each junction; releasing it with the manual deadman (LB)
held gives the operator direct driving on the analog sticks, an
instantaneous human-takeover path; releasing everything stops the vehicle.
Dataset recording timestamps every frame and command for the offline
analyses of Sec.~\ref{sec:characterization}.

% =============================================================================
\section{Experimental Characterization}
\label{sec:characterization}
\begin{figure}[t]
  \centering
  \includegraphics[width=0.82\linewidth]{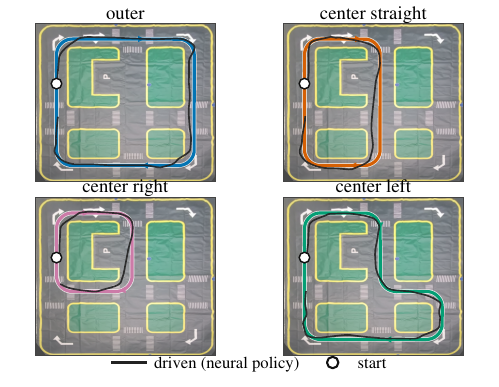}
  \caption{The printed $3.2{\times}2.8\unit{m}$ track and the four
  evaluation routes (color), with the trajectories driven by the deployed
  neural policy (dark; mean route error $3.9$--$6.5\unit{cm}$). The white
  dot marks the shared start pose.}
  \label{fig:track}
\end{figure}

% DeepPicar: measured, honest, each subsection = one question.
% =============================================================================
All measurements come from the platform's recorded sessions: twelve
closed-loop autonomous driving sessions ($5{,}760$ frames, $727\unit{s}$ of
operation) driven by convolutional policies on a printed
$3.2{\times}2.8\unit{m}$ urban track over four routes
(Fig.~\ref{fig:track}), plus the teleoperated data-collection sessions of
the same deployment~\cite{couto2026platform}. The deployment ran on a
dedicated $2.4\unit{GHz}$ 802.11n network (fixed channel~1 at
$20\unit{MHz}$ width, as recorded for the timing runs) on the consumer
router of Table~\ref{tab:bom}, with both microcontrollers under the
ESP-IDF default $20\unit{dBm}$ transmit-power cap.

\subsection{Camera link latency}
Measuring this pipeline end to end means comparing timestamps taken on three
independent monotonic clocks---camera node, chassis, and host. We synchronize
them with an NTP-style exchange filtered by minimum round-trip time and fitted
with an affine term that absorbs oscillator drift, carrying the resulting
uncertainty through every reported figure ($\pm0.6\unit{ms}$ for the camera,
$\pm1.9\unit{ms}$ for the chassis). Every frame is stamped at capture, at each
host stage, and at servo and pulse-width-modulation (PWM) application.

The camera transport delay---capture timestamp on the vision node to
arrival at the host---has a median of $27\unit{ms}$ and a 95th percentile of
$31\unit{ms}$; it is the budget's
most reproducible stage, with four closed-loop sessions totaling $2{,}994$
frames agreeing on that median within $0.1\unit{ms}$. The delay is
dominated by sensor readout and Wi-Fi transport, not by decoding: JPEG
decoding of the same frames takes $0.8\unit{ms}$ at the median. As an
independent check, we let the camera observe a millisecond chronometer on a
host display over $7{,}377$ frames: read optically, the same path measures
$65\unit{ms}$ using only the host clock. The ${\approx}37\unit{ms}$ gap contains the
display's render-to-photon pipeline and the sensor's exposure and readout,
which the capture timestamp cannot see; we read it as an upper bound on those
terms---ruling out a gross error in the capture timestamp without resolving
them.

Table~\ref{tab:budget} decomposes the full capture-to-actuation budget over
the $528$ closed-loop neural-control frames with complete capture-to-PWM
timestamp traces. Inference occupies $2.5\unit{ms}$ of a $59.5\unit{ms}$ median, confirming that
on a microcontroller-class platform the transport, not the computation, sets
the reaction time; DeepPicar reports a comparable combined
camera-plus-actuation latency of $\sim$$50\unit{ms}$ on a wired on-board
pipeline~\cite{bechtel2018deeppicar}, which our wireless split matches within
$10\unit{ms}$. Two stages deserve note. Host-side queueing between decode and
inference costs $13\unit{ms}$ at the median---more than five times inference
itself---and is scheduling slack rather than work. And the command transport
back to the chassis, while cheap at the median ($3.4\unit{ms}$), carries the
budget's heaviest tail ($65.6\unit{ms}$ at p95): it is the Wi-Fi link, not any
computation, that makes the worst case.

A FreeRTOS runtime-stats capture prices the partition: the $100\unit{Hz}$
control task uses $1\%$ of processor time, while best-effort work leaves
core~0 just $6.6\%$ idle against core~1's $25.7\%$. The loop that must not miss
its period is the cheapest task in the system, insulated by both priority and
core affinity. That core~0 figure is poll-dominated---the executor spins on a
sub-tick busy-wait---rather than a headroom cliff, and the capture is
stationary, on the instrumented build.

\begin{table}[t]
  \centering
  \caption{Capture-to-actuation latency measured during closed-loop
  neural-control operation ($528$ frames; all values in ms).}
  \label{tab:budget}
  \setlength{\tabcolsep}{4pt}
  \begin{tabular}{lrrrr}
    \toprule
    Stage & Min & Median & p95 & Max \\
    \midrule
    Camera link (capture $\rightarrow$ host)   & $26.9$ & $27.4$ & $30.6$ & $44.1$ \\
    JPEG decode                                & $0.6$ & $0.8$  & $1.1$ & $2.2$ \\
    Host queueing (decode $\rightarrow$ inference) & $0.5$ & $13.0$ & $26.3$ & $74.8$ \\
    Policy inference                           & $2.0$ & $2.5$  & $3.9$ & $5.2$ \\
    Command transport (host $\rightarrow$ chassis) & $1.3$ & $3.4$  & $65.6$ & $234.6$ \\
    Chassis (command $\rightarrow$ motor PWM)  & $1.0$ & $7.4$  & $9.6$ & $45.0$ \\
    \midrule
    End-to-end (capture $\rightarrow$ PWM)     & $35.0$ & $59.5$ & $118.7$ & $308.1$ \\
    \bottomrule
  \end{tabular}

  \smallskip
  \parbox{0.9\linewidth}{\footnotesize Stage statistics are computed
  independently and need not sum to the end-to-end row (statistics of the per-frame totals).}
\end{table}

\subsection{Effective pipeline throughput}
The recording pipeline is configured for $10\unit{Hz}$, yet the effective
labeled-frame rate observed across all twelve autonomous sessions is
consistently $7.5$--$7.9\unit{Hz}$: the wireless
hop and the host-side frame selection silently shed $20$--$25\%$ of the
nominal rate (command-acknowledgement loss measured on the same radio is
${\approx}1\%$, so frame selection, not packet loss, dominates). Worst-case inter-frame gaps reach $0.4$--$1.04\unit{s}$ in
individual sessions. Two design consequences follow: downstream consumers
must use timestamps rather than assume a fixed $10\unit{Hz}$ step, and the
gaps place a floor on the watchdog period---at $500\unit{ms}$ it triggered
during the longest observed gap, coasting the vehicle safely until frames
resumed.

\subsection{On-board odometry accuracy}
State estimation must work from what the two microcontrollers can sense:
wheel encoders, chassis geometry, and the IMU. We compared four estimators
over the same recorded signals of ten loop-shaped sessions (mean path
$7.9\unit{m}$), scoring each by loop-closure error---the distance between the
trajectory's endpoint and its start:
the firmware's encoder-only integration ($1.36\unit{m}$ mean closure), a
rear-wheel differential-speed variant ($1.35\unit{m}$), a kinematic
Ackermann model using the commanded steering angle, the only signal a
feedback-less hobby servo offers ($1.33\unit{m}$), and an
EKF fusing encoder translation with gyroscope heading ($1.06\unit{m}$, a
$22\%$ reduction; worst case $1.77\unit{m}$ against $2.66$--$2.94\unit{m}$
for the alternatives). The three encoder-based variants agree closely,
isolating \emph{heading} as the dominant drift source and explaining why the
gyroscope is the most valuable complementary signal. Even fused,
$\sim$$13\%$-of-path drift confirms that odometry alone cannot provide
evaluation-grade trajectories, motivating the platform's map-registration
pipeline~\cite{couto2026platform}.

\subsection{Closed-loop neural operation}
Under neural control the pipeline sustained continuous operation with the
policy engaged for $76$--$99\%$ of frames per session (the remainder being
deadman-off periods at session boundaries). The command-gating layer logged
every degradation event: camera timeouts occurred in eight frames across
three sessions, all absorbed by holding the last safe command; no session
ended in collision or required manual intervention. Downstream, the same deployment
achieves $5$--$7\unit{cm}$ route-following error across its
sessions~\cite{couto2026platform} ($3.9$--$6.5\unit{cm}$ on the four runs
of Fig.~\ref{fig:track}):
the measured envelope, though below nominal, suffices at the operated
$0.15\unit{m/s}$.

\subsection{Latency sensitivity in the digital twin}
\begin{figure}[t]
  \centering
  \includegraphics[width=0.82\linewidth]{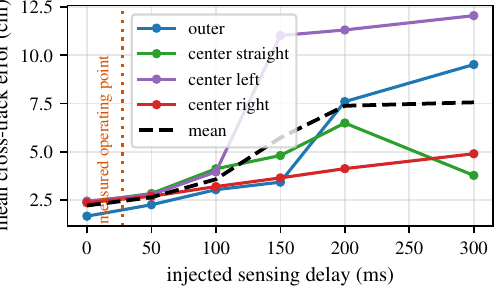}
  \caption{Mean cross-track error in the digital twin as a function of
  injected sensing delay, for the deployed policy on the four routes. The
  dotted line marks the measured median camera-link delay.}
  \label{fig:delaysweep}
\end{figure}

To relate the measured latency to control quality directly, we injected
artificial sensing delay into the platform's Webots digital
twin~\cite{couto2026platform}: the closed-loop policy sees the camera frame
from $t-\Delta$ (a frame ring buffer at simulation-step granularity), with
$\Delta$ swept from $0$ to $300\unit{ms}$ over the track's four routes. The
policy under test is the same network deployed on the physical vehicle; in
the twin, rendered frames pass through the platform's sim-to-real image
translator before inference, so the policy operates in the appearance domain
it was trained on. Fig.~\ref{fig:delaysweep} shows the resulting cross-track
error. At zero delay the policy completes all four routes at
$2.2\unit{cm}$ mean error, and degradation is graceful across the entire
sweep: no run leaves the route even at $300\unit{ms}$, $11\times$ the
measured link latency. The measured real operating point ($27\unit{ms}$
median) costs roughly $+10\%$ error relative to zero delay (about
$2.4\unit{cm}$ mean), consistent with the successful physical runs. The
most latency-sensitive route (center-left, whose two consecutive turns
amplify late corrections) grows from $2.5$ to $11$--$12\unit{cm}$ beyond
$150\unit{ms}$, while the remaining routes stay below $10\unit{cm}$
throughout. For reference, the measured $27\unit{ms}$
median sits well inside the sub-$100\unit{ms}$ envelope that commercial
teledriving targets over bonded cellular links~\cite{neumeier2019teleop,kamtam2024teleop}, and the
largest delay tested remains well below human driver perception--brake
times of $0.7$--$1.5\unit{s}$~\cite{green2000stop}. Two caveats bound how far
these margins carry. The twin injects a constant delay, whereas the real
budget is heavy-tailed, and that tail belongs to the radio rather than to the
platform: the p95 of $119\unit{ms}$ in Table~\ref{tab:budget} is twice its
median, and three sessions minutes apart put it near $80\unit{ms}$. A constant
delay is not the same regime as a bursty one, so the
margin should be read against the tail rather than the median. The twin also
injects delay on the sensing path only; the command path, whose tail is the
budget's heaviest (Table~\ref{tab:budget}), is not swept. And delay harms
only through the distance covered while it elapses: at the $0.15\unit{m/s}$
operated here, $300\unit{ms}$ is under $5\unit{cm}$ of travel, so these
margins are a statement about this speed, not a speed-free property of the
policy.

\subsection{Speed-loop step response}
\begin{figure}[t]
  \centering
  \includegraphics[width=0.82\linewidth]{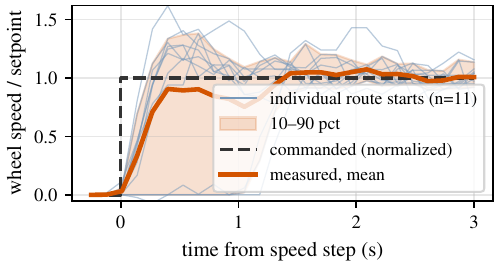}
  \caption{Normalized speed-loop step response at autonomous-session
  starts. Thin lines: individual steps; thick line: their mean; dashed: the
  commanded step; band: 10--90th percentile.}
  \label{fig:pidstep}
\end{figure}

The closed-loop behavior of the $100\unit{Hz}$ PID speed loop can be read
from the recorded sessions themselves: every deadman engagement is a speed
step from rest, observed through the commanded and measured wheel speeds in
the telemetry. Fig.~\ref{fig:pidstep} aggregates the eleven such steps in the
autonomous sessions, one per qualifying session, normalized by the setpoint. The loop reaches $90\%$ of
the setpoint in about $0.5\unit{s}$ with negligible mean overshoot; one run
starts $\sim$$1\unit{s}$ late, attributed to static friction at standstill. Rise time is not setpoint-free, however: a dedicated bench step
test at $20\unit{Hz}$ reaches $90\%$ in $0.43\unit{s}$ at $0.10\unit{m/s}$ but
$0.98\unit{s}$ at $0.15$ and $1.58\unit{s}$ at $0.05$---slowest where
quantization is coarsest---with steady-state error within $\pm4\%$ throughout.
The salient systems finding is the sustained speed ripple of
roughly $\pm20\%$ of the setpoint at the operated $0.135\unit{m/s}$: at this
speed the $836$-count encoders yield few counts per $10\unit{ms}$ control
period, so measurement quantization---not the controller---bounds the speed
regulation quality. That same test makes the quantum explicit: the reported
wheel speed takes only fifteen distinct values, each an exact multiple of
$12.87\unit{mm/s}$---$8.6\%$ of the setpoint at $0.15\unit{m/s}$ but $25.7\%$
at $0.05$---which accounts for the ripple and predicts its growth as speed
falls. Vehicle inertia filters the ripple---route following is
unaffected---but it marks the encoder resolution as the binding constraint
for lower-speed operation.

% =============================================================================
\section{Conclusion}
\label{sec:conclusion}
% =============================================================================
\begin{table}[t]
  \centering
  \caption{Design lessons from the characterization.}
  \label{tab:lessons}
  \scriptsize
  \begin{tabular}{@{}p{0.26\linewidth}p{0.29\linewidth}p{0.33\linewidth}@{}}
    \toprule
    Finding & Evidence & Design implication \\
    \midrule
    Wireless link dominates latency & camera link $27\unit{ms}$ vs.\
    $2.5\unit{ms}$ inference & optimize transport before the network \\
    \addlinespace[2pt]
    End-to-end timing has heavy tails & p95 twice the median; gaps up to
    $1.04\unit{s}$ & timestamped processing; watchdog protection \\
    \addlinespace[2pt]
    Heading dominates odometry drift & EKF fusion cuts loop-closure error
    $22\%$ & gyroscope fusion pays even at this scale \\
    \addlinespace[2pt]
    Encoder quantization limits low-speed PID & sustained $\pm20\%$ speed
    ripple & higher-resolution encoders for lower speeds \\
    \bottomrule
  \end{tabular}
\end{table}

We presented the embedded architecture and real-time characterization of a
low-cost autonomous research vehicle organized as a perceive--decide--act
agent: perception isolated on an ESP32 camera node, decision offloaded to a
host GPU over Wi-Fi, and action---all safety-relevant control---in a
$100\unit{Hz}$ loop on an ESP32-S3 under micro-ROS. Twelve real autonomous
sessions support the four systems lessons of Table~\ref{tab:lessons}. The
partition holds: the loop that must not miss its period is the system's
cheapest task. Within this envelope the platform sustains
centimeter-level route-following---evidence that deliberately partitioned
microcontroller-class vehicles are sufficient testbeds for embedded neural
driving research, operated as teledriving-inspired supervised autonomy in
miniature: the policy drives under the operator's guidance and
instant-takeover grip. The wireless offload mirrors commercial
teledriving over bonded cellular links engineered for sub-$100\unit{ms}$
glass-to-glass latency~\cite{neumeier2019teleop,kamtam2024teleop}; the measured
budget places the platform in that regime. Future work will finish the
on-MCU characterization, replicate the digital-twin delay injection on the
physical vehicle, and extend operation beyond line of sight, toward forest
surveillance~\cite{oliveira2021forest} and disaster-site
inspection~\cite{delmerico2019rescue}.

\section*{Acknowledgment}
This work was supported by the National Council for Scientific and
Technological Development -- CNPq, Brazil (Grant No.\ 420148/2025-6), and
financed in part by the Coordena\c{c}\~ao de Aperfei\c{c}oamento de Pessoal
de N\'ivel Superior -- Brasil (CAPES) -- Finance Code 001.

\bibliographystyle{IEEEtran}
\bibliography{references}

@inproceedings{bechtel2018deeppicar,
  title={{DeepPicar}: A Low-Cost Deep Neural Network-Based Autonomous Car},
  author={Bechtel, Michael G. and McEllhiney, Elise and Kim, Minje and Yun, Heechul},
  booktitle={IEEE International Conference on Embedded and Real-Time Computing Systems and Applications (RTCSA)},
  pages={11--21},
  year={2018}
}

@inproceedings{okelly2020f1tenth,
  title={{F1TENTH}: An Open-source Evaluation Environment for Continuous Control and Reinforcement Learning},
  author={O'Kelly, Matthew and Zheng, Hongrui and Karthik, Dhruv and Mangharam, Rahul},
  booktitle={NeurIPS 2019 Competition and Demonstration Track},
  year={2020}
}

@inproceedings{paull2017duckietown,
  title={Duckietown: An Open, Inexpensive and Flexible Platform for Autonomy Education and Research},
  author={Paull, Liam and Tani, Jacopo and Ahn, Heejin and Alonso-Mora, Javier and Carlone, Luca and Cap, Michal and Chen, Yu Fan and Choi, Changhyun and Dusek, Jeff and Fang, Yajun and others},
  booktitle={IEEE International Conference on Robotics and Automation (ICRA)},
  pages={1497--1504},
  year={2017}
}

@article{belsare2023micro,
  title={Micro-{ROS}},
  author={Belsare, Kaiwalya and others},
  journal={Robot Operating System (ROS): The Complete Reference (Volume 7)},
  pages={3--55},
  publisher={Springer},
  year={2023}
}

@inproceedings{moore2016generalized,
  title={A Generalized Extended {Kalman} Filter Implementation for the Robot Operating System},
  author={Moore, Thomas and Stouch, Daniel},
  booktitle={Intelligent Autonomous Systems 13 (IAS-13)},
  pages={335--348},
  publisher={Springer},
  year={2016}
}

@article{zhang2000flexible,
  title={A Flexible New Technique for Camera Calibration},
  author={Zhang, Zhengyou},
  journal={IEEE Transactions on Pattern Analysis and Machine Intelligence},
  volume={22},
  number={11},
  pages={1330--1334},
  year={2000}
}

@article{bojarski2016,
  title={End to End Learning for Self-Driving Cars},
  author={Bojarski, Mariusz and Del Testa, Davide and Dworakowski, Daniel and Firner, Bernhard and Flepp, Beat and Goyal, Prasoon and Jackel, Lawrence D. and Monfort, Mathew and Muller, Urs and Zhang, Jiakai and others},
  journal={arXiv preprint arXiv:1604.07316},
  year={2016}
}

@article{couto2023hgail,
  title={Hierarchical Generative Adversarial Imitation Learning with Mid-Level Input Generation for Autonomous Driving on Urban Environments},
  author={Couto, Gustavo Claudio Karl and Antonelo, Eric Aislan},
  journal={IEEE Transactions on Intelligent Vehicles},
  year={2024}
}

@inproceedings{couto2026platform,
  title={A Low-Cost, Open Platform for End-to-End Autonomous Driving on a Miniature {Ackermann} Vehicle},
  author={Couto, Gustavo Claudio Karl and Antonelo, Eric Aislan and Zipperer, Gabriel George},
  booktitle={Proceedings of the XXIII Encontro Nacional de Intelig{\^e}ncia Artificial e Computacional (ENIAC)},
  publisher={SBC},
  year={2026},
  note={Accepted; to appear. Preprint: arXiv:2609.04147}
}

@article{burns2017mixed,
  title={A survey of research into mixed criticality systems},
  author={Burns, Alan and Davis, Robert I.},
  journal={ACM Computing Surveys},
  volume={50},
  number={6},
  pages={82:1--82:37},
  year={2017}
}

@article{sprunt1989aperiodic,
  title={Aperiodic task scheduling for hard-real-time systems},
  author={Sprunt, Brinkley and Sha, Lui and Lehoczky, John},
  journal={Real-Time Systems},
  volume={1},
  number={1},
  pages={27--60},
  year={1989}
}

@article{kamtam2024teleop,
  author  = {Kamtam, Sidharth Bhanu and Lu, Qian and Bouali, Faouzi and Haas, Olivier C. L. and Birrell, Stewart},
  title   = {Network Latency in Teleoperation of Connected and Autonomous Vehicles: A Review of Trends, Challenges, and Mitigation Strategies},
  journal = {Sensors},
  volume  = {24},
  number  = {12},
  pages   = {3957},
  year    = {2024}
}

@article{green2000stop,
  author  = {Green, Marc},
  title   = {``How Long Does It Take to Stop?'' {M}ethodological Analysis of Driver Perception-Brake Times},
  journal = {Transportation Human Factors},
  volume  = {2},
  number  = {3},
  pages   = {195--216},
  year    = {2000}
}

@article{oliveira2021forest,
  author  = {Oliveira, Lu{\'i}s F. P. and Moreira, Ant{\'o}nio P. and Silva, Manuel F.},
  title   = {Advances in Forest Robotics: A State-of-the-Art Survey},
  journal = {Robotics},
  volume  = {10},
  number  = {2},
  pages   = {53},
  year    = {2021}
}

@article{delmerico2019rescue,
  author  = {Delmerico, Jeffrey and Mintchev, Stefano and Giusti, Alessandro and others},
  title   = {The Current State and Future Outlook of Rescue Robotics},
  journal = {Journal of Field Robotics},
  volume  = {36},
  number  = {7},
  pages   = {1171--1191},
  year    = {2019}
}

@inproceedings{neumeier2019teleop,
  author    = {Neumeier, Stefan and Wintersberger, Philipp and Frison, Anna-Katharina and Becher, Armin and Facchi, Christian and Riener, Andreas},
  title     = {Teleoperation: The Holy Grail to Solve Problems of Automated Driving? {S}ure, but Latency Matters},
  booktitle = {Proceedings of the 11th International Conference on Automotive User Interfaces and Interactive Vehicular Applications (AutomotiveUI)},
  year      = {2019},
  pages     = {186--197},
  publisher = {ACM}
}

\end{document}